%% file: arxiv.tex
\pdfoutput=1
\documentclass[11pt]{article}

\usepackage[final]{acl}

\usepackage{tcolorbox}
\usepackage{caption}
\usepackage{geometry}
\usepackage{times}
\usepackage{latexsym}
\usepackage{colortbl, xcolor} 
\usepackage{times}
\usepackage{latexsym}
\usepackage{adjustbox}
\usepackage[T1]{fontenc}
\usepackage[utf8]{inputenc}
\usepackage{microtype}
\usepackage{inconsolata}
\usepackage{graphicx}
\usepackage{amsmath}
\usepackage{mathrsfs}
\usepackage{multirow}
\usepackage{multicol}
\usepackage{booktabs}
\usepackage{amsmath}
\usepackage{algorithm}
\usepackage{algpseudocode}

\usepackage{newunicodechar}

\definecolor{lightgray}{gray}{0.9}
\definecolor{headergray}{gray}{0.8}
\definecolor{darkgray}{gray}{0.6}
\usepackage{makecell}
\newcommand\blfootnote[1]{%
  \begingroup
  \renewcommand\thefootnote{}\footnote{#1}%
  \addtocounter{footnote}{-1}%
  \endgroup
}

\title{KG-CQR: Leveraging Structured Relation Representations in Knowledge Graphs for Contextual Query Retrieval}

\author{
 \textbf{Chi Minh Bui\textsuperscript{1}$^\ast$},
 \textbf{Ngoc Mai Thieu\textsuperscript{1}$^\ast$},
 \textbf{Van Vinh Nguyen\textsuperscript{2}},
 \textbf{Jason J. Jung\textsuperscript{3}},
    \textbf{Khac-Hoai Nam Bui\textsuperscript{1}$^\dag$}
\\
\textsuperscript{1}Viettel AI, Viettel Group, Vietnam \\
\textsuperscript{2}University of Engineering and Technology, Vietnam National University, Hanoi, Vietnam \\
 \textsuperscript{3}Department of Computer Engineering, Chung-Ang University, Korea
\\
\small{
\{minhbc4, maitn4\}@viettel.com.vn, vinhvn@vnu.edu.vn, j2jung@gmail.com, nambkh@viettel.com.vn}
}

\begin{document}
\maketitle
\input{sections/0_Abstract}
\blfootnote{$^\ast$ Equal Contribution}
\blfootnote{$^\dag$ Corresponding Author}
\input{sections/01_Introduction}
\input{sections/02_Literature_review}
\input{sections/03_Methodology}
\input{sections/04_Experiment}
\input{sections/05_Conclusion}

\input{sections/Limitations}

\bibliography{custom}
\input{sections/06_Appendix}

\end{document}

%% file: sections/0_Abstract.tex
\begin{abstract}
The integration of knowledge graphs (KGs) with large language models (LLMs) offers significant potential to enhance the retrieval stage in retrieval-augmented generation (RAG) systems. In this study, we propose KG-CQR\footnote{https://github.com/tnmai59/KG-CQR}, a novel framework for Contextual Query Retrieval (CQR) that enhances the retrieval phase by enriching complex input queries with contextual representations derived from a corpus-centric KG. Unlike existing methods that primarily address corpus-level context loss, KG-CQR focuses on query enrichment through structured relation representations, extracting and completing relevant KG subgraphs to generate semantically rich query contexts. Comprising subgraph extraction, completion, and contextual generation modules, KG-CQR operates as a model-agnostic pipeline, ensuring scalability across LLMs of varying sizes without additional training. Experimental results on the RAGBench and MultiHop-RAG datasets demonstrate that KG-CQR outperforms strong baselines, achieving improvements of up to 4–6\% in mAP and approximately 2–3\% in Recall@25. Furthermore, evaluations on challenging RAG tasks such as multi-hop question answering show that, by incorporating KG-CQR, the performance outperforms the existing baseline in terms of retrieval effectiveness.
\end{abstract}

%% file: sections/01_Introduction.tex
\section{Introduction}
Large Language Models (LLMs) have significantly advanced the field of natural language processing (NLP), particularly in understanding and generating human-like text. However, LLMs still suffer from two critical limitations: a lack of reliable factual knowledge and limited reasoning capabilities \cite{wang-etal-2024-factuality}. 
\begin{figure}[!ht]
    \centering
    \includegraphics[width=\linewidth]{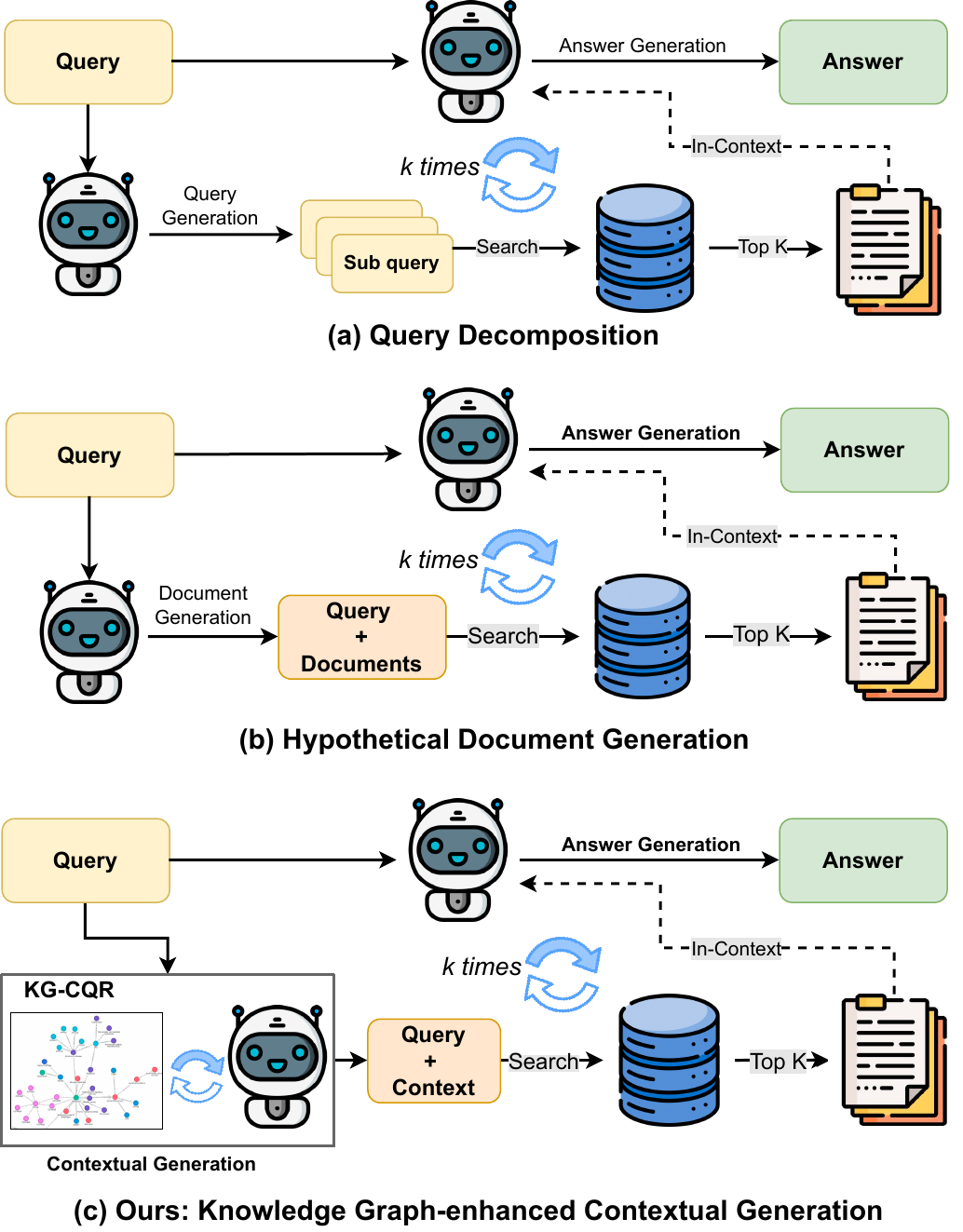}
    \caption{Overview of query expansion approaches for RAG systems: a) query decomposition; b) document generation; c) ours: KG-enhanced contextual generation}
    \label{fig1}
\end{figure} 
These limitations are exacerbated when LLMs are applied to domain-specific knowledge retrieval, especially in addressing queries within vertical domains \cite{BangCLDSWLJYCDXF23}.
To address these challenges, recent research has explored the integration of knowledge graphs (KGs) into LLMs as a means to provide structured, accurate knowledge sources for enhanced reasoning \cite{PanLWCWW24}. KGs, which store facts in the form of triples (i.e., head entity, relation, tail entity), offer a robust and interpretable representation of knowledge. Consequently, KGs have been increasingly incorporated into applications based on LLMs to improve performance across various tasks, such as question answering \cite{DingLLQ24}, fact verification \cite{pham-etal-2025-claimpkg}, and recommendation systems \cite{Abu-Rasheed0F24}.

In the context of question answering over knowledge graphs (KGQA), current approaches can be broadly categorized into two main strategies: (i) using LLMs to convert natural language queries into formal logical queries, which are then executed on KGs to derive answers \cite{NguyenLS0LVH24, abs-2404-10384}; and (ii) retrieving relevant triples from KGs and presenting them as contextual knowledge for the LLM to generate the final answer \cite{SarmahMHRPP24, SunXTW0GNSG24}. Similarly, in retrieval-augmented generation (RAG) tasks, external knowledge sources, in terms of both structured (KGs) and unstructured (vectorized documents), are retrieved and incorporated into the input prompt to support answer generation by LLMs \cite{LiZCDJPB24, abs-2404-16130}. Despite these advances, the retrieval process involving KGs remains underexplored in the aforementioned approaches. 

This study focuses on enhancing the retrieval process for RAG systems by integrating KG technologies to enrich contextual information for complex input queries.
Specifically, the objective is to tackle a critical challenge in current systems: misalignment between query and document embeddings \cite{ma-etal-2023-query}. Accordingly, existing methods often employ LLMs to decompose complex queries \cite{mao-etal-2024-rafe} (Figure \ref{fig1}(a)). Nonetheless, in terms of retrieval performance, this approach frequently underperforms due to insufficient contextual alignment with the corpus. Subsequently, \citet{GaoMLC23} proposed a new approach by generating hypothetical documents to facilitate document-document similarity comparisons (Figure \ref{fig1}(b)). However, this method heavily relies on underlying LLMs, introducing risks of hallucination. In terms of knowledge-grounded expansion generation, \citet{xia-etal-2025-knowledge} introduced a knowledge-aware approach that leverages both unstructured data and structured relations. Nevertheless, their reliance on predefined relation schemas between entities (e.g., title) and documents constrains the scalability and adaptability. 

To overcome the aforementioned limitations, we propose KG-CQR (Knowledge Graph for Contextual Query Retrieval), a novel framework that leverages KG to generate contextual information for input queries (Figure \ref{fig1}(c)). The key idea is to extract a relevant subgraph from the KG to enrich each query semantically. KG-CQR comprises three main modules: (i) subgraph extraction, which identifies relevant triples; (ii) subgraph completion, which infers missing triples; and (iii) contextual generation, which constructs enriched query contexts. These modules utilize a new structured representation of relations, combining textual information with KG triplets, to address the limitations of traditional entity-based scoring in KG extraction. By retrieving directly relevant data and inferring missing knowledge, KG-CQR significantly improves query contextualization. The main contributions of this work are as follows:
\begin{itemize}
    \item We propose Contextual Query Retrieval (CQR), a novel paradigm designed to enhance the context of domain-specific queries using a predefined corpus. Our framework, KG-CQR, leverages a corpus-centric knowledge graph to improve both query understanding and retrieval effectiveness, achieving these improvements without the need for additional training.
    \item KG-CQR functions as a model-agnostic pipeline that employs structured relation representations to generate contextual information, ensuring adaptability and scalability across backbone LLMs with varying parameter sizes.
    \item Extensive experiments on complex benchmark datasets, specifically designed for multi-step retrieval processes in RAG systems. The results demonstrate the effectiveness of KG-CQR in enhancing retrieval quality.
\end{itemize} 

%% file: sections/02_Literature_review.tex
\section{Literature Review}
\subsection{Query Expansion using LLM}
To handle complex queries effectively, query expansion is often essential for improving the performance of the retrieval process \cite{AzadD19}. Traditional approaches decompose input queries into multi-view representations to enhance retrieval accuracy \cite{ZhangLGJD22}. Recently, with the rapid advancement of LLMs, a promising direction involves query enhancement, either through prompt-based techniques leveraging LLMs \cite{WangYW23}, or by developing trainable frameworks that generate refined queries \cite{mao-etal-2024-rafe}. These methods aim to reformulate queries into more effective semantic representations \cite{abs-2404-00610, ChenCHW024}. However, they still struggle to bridge the inherent gap between queries and the knowledge corpus within the retrieval embedding space \cite{LiuPCBSZCWYD25}. Accordingly, to further improve retrieval effectiveness, especially in domain-specific applications, a deeper exploitation of contextual generation remains essential \cite{10.1145/3722552}.

\subsection{Contextual Retrieval}
Contextualized retrieval has emerged as an effective strategy for improving retrieval performance, particularly in complex and challenging settings \cite{abs-2410-02525}. Recent methods, such as RAPTOR \cite{SarthiATKGM24}, GraphRAG \cite{abs-2404-16130}, and HippoRAG \cite{GutierrezS0Y024}, adopt recursive procedures that integrate embedding, clustering, and summarization techniques to construct hierarchical document representations using graph-based structures. Conceptually, these approaches follow a corpus-centric paradigm, wherein hierarchical structures are leveraged to enhance contextual retrieval across the original document corpus. In terms of query expansion through contextualization, \citet{GaoMLC23} proposes HyDE, a novel approach that leverages LLMs to generate hypothetical documents conditioned on the input query. Accordingly, the query is first processed by an LLM following specific instructions to produce hypothetical documents, which are then used as pseudo-contexts for retrieval based on document-to-document similarity. However, a key limitation of HyDE lies in its dependence on LLM-generated content, where potential inaccuracies or hallucinations can degrade retrieval effectiveness \cite{ZhangWYN24, xia-etal-2025-knowledge}. Moreover, query expansion strategies must account for domain-specific context sensitivity, as the same entities may vary in meaning or relevance across different domains \cite{BuiTTP21}.
Therefore, this study proposes a novel contextual retrieval approach, which focuses on providing contextual information for the input query, based on the structured relation of the corpus-centric KG. 

\subsection{LLM-Powered KG Construction}
One of the primary challenges in utilizing knowledge graphs (KGs) lies in their construction. Prior work relies on predefined KGs \cite{xia-etal-2025-knowledge}, which limits the flexibility and scalability of the approach. In order to automatically construct a KG, given a set of unstructured data sources (corpus), knowledge graph construction (KGC) is typically framed as a structured prediction task, where models are trained to approximate target functions associated with various NLP tasks such as Named Entity Recognition (NER), Relation Extraction (RE), Entity Linking (EL), and knowledge graph completion \cite{Ye0CC22}. However, training task-specific discriminative models often results in error propagation and limited adaptability across diverse tasks. To address these limitations, recent approaches reformulate KGC as a generative problem using sequence-to-sequence (Seq2Seq) models \cite{0001LDXLHSW22}. Powered by pre-trained models such as T5 \cite{RaffelSRLNMZLL20}, the Seq2Seq paradigm has demonstrated strong performance in multi-task training settings for KG construction. More recently, the emergence of LLMs has spurred interest in their application to KGC through zero-shot prompting techniques \cite{PanLWCWW24, ZhuWCQOYDCZ24}. Building on this line of work, our study leverages modern open-source LLMs, e.g., LLaMA-3.3-70B, to construct knowledge graphs by parsing and categorizing entities and their relationships directly from unstructured data.

%% file: sections/03_Methodology.tex
\section{Methodology}
\subsection{Preliminary}
\label{preliminary}
\subsubsection{Structure Relation Representation}
\label{KG-construction}
A corpus-centric KG includes a set of triplets (structured relations) $\mathcal{T}_{KG}$, which are defined as follows:
\begin{equation}
\begin{aligned}
     KG = \{E_{KG}, R_{KG}, \mathcal{T}_{KG}\}
      \\
    \mathcal{T}_{KG} = \{(u, r, v), u, v \in E_{KG}, r \in R_{KG}\}
\end{aligned}
\label{SRR}
\end{equation}
where $E_{KG}$ is the set of entities and $R_{KG}$ is the set of relations. Since the KG is not available for most specific domains, we follow the work in GraphRAG \cite {abs-2404-16130} to construct the corpus-centric KG, which includes three sequential steps: i) Ingesting specific-domain unstructured data; ii) Extracting entities and their relationships using an external LLM; iii) Mapping entities through edges (relations) that contain detailed information about their relationships. 
\begin{figure}[!ht]
    \centering
    \includegraphics[width=\linewidth]{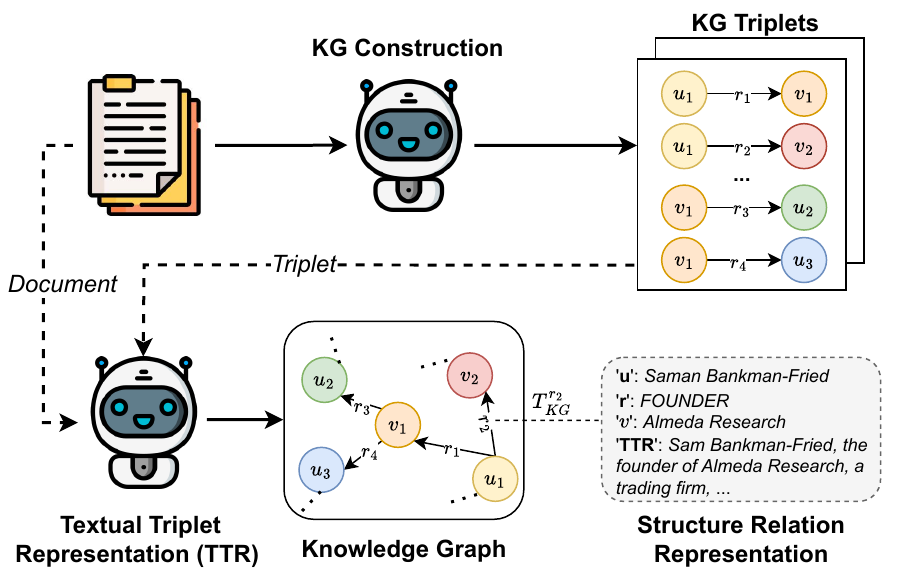}
    \caption{Construction of structured relation representations using LLM-based prompting. Detailed prompt templates are provided in Appendix \ref{prompt-template}.}
    \label{fig2}
\end{figure}

To further enhance the expressiveness of the KG, we extend each triplet $\mathcal{T}_{KG}^i$ with a textual triplet representation (TTR). Unlike traditional approaches that rely solely on structured relational properties, our method leverages LLMs to generate rich, natural language representations of each triplet, as defined below:
\begin{equation}
\begin{aligned}
    TTR(\mathcal{T}_{KG}^i) = llm(Prompt_{ttr}, D_d^i, \mathcal{T}_{KG}^i)
\end{aligned}
\label{triplet_embeddings}
\end{equation}
where $\text{llm}(Prompt_{ttr}, D_d^i, \mathcal{T}_{KG}^i)$ denotes the textual description of the relation, generated by an LLM based on the instruction prompt $Prompt_{ttr}$, the corresponding triplet $\mathcal{T}_{KG}^i$, and the document $d \in D$ from which the triplet was extracted. An overview of this process is illustrated in Figure \ref{fig2}.
In this regard, the structured relation in Equation \ref{SRR} is reformulated as:
\begin{equation}
\begin{aligned}
  \mathcal{T}_{KG} = \{(u, r, v, TTR(u,r,v))\}
\end{aligned}
\end{equation}

\subsubsection{Problem Definition}
The objective of the retrieval process is to extract the most relevant documents for the input query, in which the similarity score (i.e., cosine similarity) can be formulated as follows:
\begin{equation}
 sim(q, d) = <\mathbf{v}_q, \mathbf{v_d}>
 \label{objectivefunction}
\end{equation}
The core challenge in this process lies in ensuring that the query vector $\mathbf{v}_q$ (obtained via encoder $enc_q$) and the document vector $\mathbf{v}_d$ (obtained via encoder $enc_d$) are embedded into a shared semantic space. Traditional retrieval models typically rely on supervised learning frameworks that train encoders using query-document pairs to learn such a shared embedding space \cite{KarpukhinOMLWEC20, SanthanamKSPZ22}.
However, directly optimizing for query-document similarity often results in suboptimal retrieval performance, particularly when dealing with sparse or domain-specific queries. To address this limitation, we draw inspiration from the approach in \cite{GaoMLC23}, which shifts focus toward generating contextual embeddings for the query. Notably, instead of encoding the query directly, we enrich it with contextual information derived from the corpus-centric KG. This enriched representation is then embedded in the document space, allowing the similarity computation to align with the document-document similarity paradigm. The revised retrieval formulation is as follows:
\begin{equation}
\begin{aligned}
\mathbf{v}_{\text{KG-CQR(q)}} = enc_d(\text{KG-CQR(q)})
\\
 sim(q,d) = 
   <\mathbf{v}_{\text{KG-CQR(q)}}, \mathbf{v}_{d}>
\end{aligned}
   \label{objective}
\end{equation}
Here, $\text{KG-CQR(q)}$ denotes the KG-enhanced contextual information of the input query $q$. 

\subsection{KG-CQR}
\begin{figure*}[!ht]
    \centering
    \includegraphics[width=\linewidth]{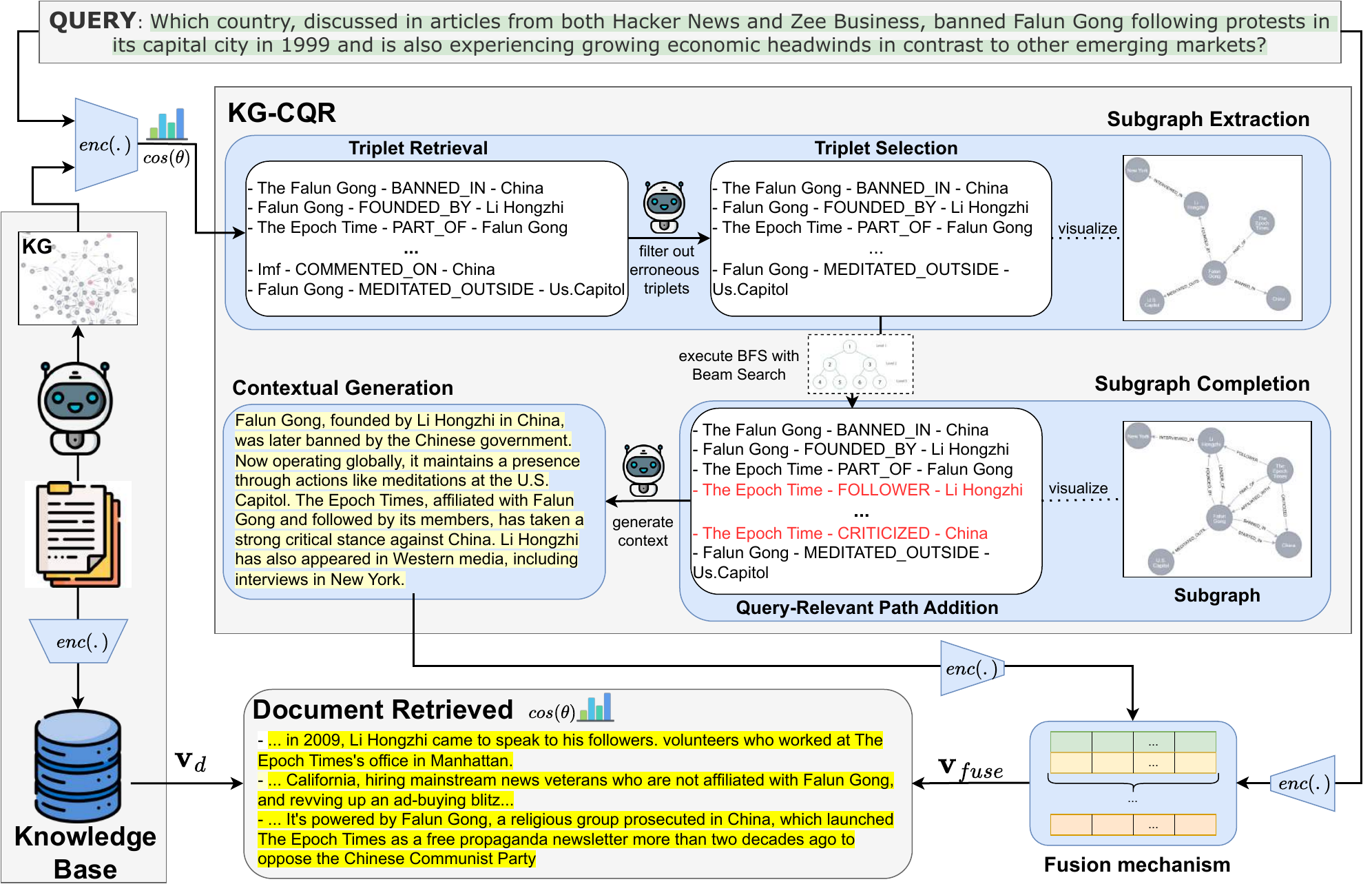}
    \caption{An illustration of KG-CQR for the retrieval process, which includes three main components: Subgraph Extraction, Subgraph Completion, and  Contextual Generation}
    \label{fig3}
\end{figure*}

The overview architecture of KG-CQR is illustrated in Figure \ref{fig3}, which includes three main sequence components, such as subgraph extraction, subgraph completion, and contextual generation.

\subsubsection{Subgraph Extraction}
Given an input query $q$ and a knowledge graph $KG$, the subgraph extraction module first identifies a set of relevant triples $\hat{\mathcal{T}}_{KG}$ ($\hat{\mathcal{T}}_{KG} \subset \mathcal{T}_{KG}$), based on the input query. Traditional subgraph extraction methods typically begin by identifying entities mentioned in the query $q$ and then linking them to entities in the KG using entity linking (EL) techniques, such as using LLM prompting or specialized EL tools \cite{SunXTW0GNSG24}. However, these approaches often assume that the KG is complete, i.e., all factual triples relevant to the query are present in the graph, which is rarely the case in real-world scenarios \cite{XuHC0STL0024}.
Furthermore, current subgraph extraction techniques predominantly rely on assessing semantic similarity at the entity or keyword level \cite{SunXTW0GNSG24, LuoLHP24}. Nevertheless, this limited granularity often fails to capture sufficient textual context, thereby reducing extraction performance, particularly when input queries involve ambiguous entities \cite{pham-etal-2025-verify, xia-etal-2025-knowledge}. To address these limitations, we leverage textual representations of triples (as defined in Equation \ref{triplet_embeddings}) to measure similarity with the input query. 
This approach enables subgraph extraction at the sentence level, rather than relying solely on the entity level. 
The subgraph extraction is formalized as follows:
\begin{equation}
\begin{aligned}
\mathbf{v}^i_{r} = enc(TTR(\mathcal{T}_{KG}^i))
 \\
 \hat{\mathcal{T}}_{KG} = \operatorname{argmax}_{\mathcal{T}^i_{KG} \in \mathcal{T}_{KG}, \ k} \{ sim(\mathbf{v}_q, \mathbf{v}^i_r)\}
\end{aligned}
\end{equation}
where $\mathbf{v}_q$ is the embedding of the input query, and $k$ is a hyperparameter controlling the number of top-matching triples retrieved.

Sequentially, inspired by previous work for the subgraph extraction process \cite{SunXTW0GNSG24}, a filtering step is performed using an LLM with a task-specific prompt to remove irrelevant triples:
\begin{equation}
\begin{aligned}
\hat{\mathcal{T'}}_{KG} = \{ \mathcal{T}_{KG}^i \in \hat{\mathcal{T}}_{KG} | 
 \\ 
 llm(Prompt_{filter}, q, \mathcal{T}_{KG}^i)  = True \}
\end{aligned}
\end{equation}
Here, $Prompt_{filter}$ denotes the instruction prompt used by the LLM for the final selection. The details of $Prompt_{filter}$ are provided in Appendix \ref{prompt-template}.

\subsubsection{Subgraph Completion}
The initial subgraph $\hat{\mathcal{T'}}_{KG}$ is extracted based on semantic similarity, typically resulting in a limited set of triplets that may lack sufficient contextual information. The goal of the subgraph completion function is to enrich this subgraph by incorporating additional triplets from the structure relation of KG ($\mathcal{T}_{KG}$) that form semantically meaningful paths between entities in $\hat{\mathcal{T'}}_{KG}$. Relevance is assessed by aggregating the semantic similarities between the input query and triplet textual representations along these paths. 
\begin{algorithm}[!ht]
\caption{Query-Relevant Path Addition for Subgraph Completion}\label{subgraph_completion}
\begin{algorithmic}[1]

\Require $\mathcal{T}_{KG}$, $\hat{\mathcal{T'}}_{KG}$, $q$, top $K$, max-path $L$

\Ensure Subgraph $\hat{\mathcal{T''}}_{KG}$

\State $E_p \gets \{u, v \mid \{u, r, v, \text{TTR}\} \in \hat{\mathcal{T'}}_{KG}\}$ 
\State Load Embedding model: enc
\State $v_q \gets q \neq \emptyset ? \text{enc}(q) : \text{None}$ 

\State $T_{\text{set}} \gets \{\{u, r, v\} \mid \{u, r, v, \text{TTR}\} \in \hat{\mathcal{T'}}_{KG}\}$ 
\State $P \gets \bigcup_{(e_i,e_j)\in E_p} \text{BFSBeam}(\mathcal{T}_{KG}, e_i, e_j, T_{\text{set}}, L)$ 
\If{$P = \emptyset$}
    \State \Return $\hat{\mathcal{T'}}_{KG}$
\EndIf

\State $S \gets \emptyset$ 

\For{$p \in P$} 
    \If{$\{v_p \gets \text{enc}(\text{TTR}) \mid \{u, r, v, \text{TTR}\} \in p\}$}
        \State $s \gets v_q \neq \text{None} ? \text{Mean}(\cos(v_p, v_q)) : 0$ 
        \State $S \gets S \cup \{(p, s)\}$ 
    \EndIf
\EndFor

\State Sort $S$ by score descending
\State $C \gets \emptyset$ 
\For{$(p, s) \in S$ until $|C| \geq K$ do} 
    \If{$\{u, r, v\} \in p \land \{u, r, v\} \notin T_{\text{set}}$}
        \State $C \gets C \cup \{u, r, v\}$ 
    \EndIf
\EndFor

\State $\hat{\mathcal{T''}}_{KG} \gets \hat{\mathcal{T'}}_{KG} \cup C$ 

\State \Return $\hat{\mathcal{T''}}_{KG}$

\end{algorithmic}
\end{algorithm}
The subgraph completion proceeds through the following steps (Algorithm~\ref{subgraph_completion}):
\begin{itemize}
\item Step 1: Extract entities from the initial subgraph $\hat{\mathcal{T'}}_{KG}$.
\item Step 2: Apply Beam Search, a heuristic-guided variant of Breadth-First Search (BFS), to identify the top-n candidate paths.
\item Step 3: Filter out paths that contain nodes not present in the initial subgraph $\hat{\mathcal{T'}}_{KG}$.
\item Step 4: Select the top-K highest-scoring unique triplets, with K defaulting to 20.
\item Step 5: Construct the completed subgraph $\hat{\mathcal{T''}}_{KG}$ by merging the initial subgraph $\hat{\mathcal{T'}}_{KG}$ with the selected triplets.
\end{itemize}
Notably, to reduce computational complexity in Step 2, instead of executing the naive BFS traversal, a limited number of nodes are expanded, guided by a heuristic function (BFSBeam). This function computes semantic similarity between the input query and aggregates the relevance scores of the TTRs along each path, which is illustrated in more detail in the Appendix~\ref{BFS-section}. 
\subsubsection{Contextual Generation}
The objective of the retrieval process is to identify the most relevant documents for a given input query by computing similarity scores, typically using cosine similarity between their vector representations, which is formally defined as:
\begin{equation}
    KG\textit{-}CQR(q) = llm(Prompt_{g}, \hat{\mathcal{T''}}_{KG})
\end{equation}
where $Prompt_{g}$ represents the generation instruction prompt, as detailed in Appendix~\ref{prompt-template}. The enriched subgraph $\hat{\mathcal{T''}}_{KG}$ serves as contextual input to the LLM, facilitating the generation of a contextually enriched query representation. This reformulated query can then be encoded within the same embedding space as the corpus documents, enabling effective retrieval.
\subsection{Retrieval Fusion Function}
The input query and its synthetic contextual information are embedded using a fusion encoder-based approach. This technique enables the retrieval system to go beyond superficial query-document matching by leveraging the interaction between the query and its enriched context, resulting in more accurate and semantically relevant retrieval outcomes \cite{BruchGI24}. In this work, we adopt a weighted-sum fusion mechanism to compute the final query representation, defined as:
\begin{equation}
\mathbf{v}_{fuse(q)} = \alpha \cdot \mathbf{v}_{q} + (1 - \alpha) \cdot \mathbf{v}_{KG-CQR(q)}
\label{alpha_fuse}
\end{equation}
This fusion mechanism proves especially effective in complex, multi-turn, or context-sensitive retrieval scenarios, where conventional query enhancement or decomposition methods often fall short. Consequently, the objective function in Equation~\ref{objective} can be reformulated as:
\begin{equation}
\begin{aligned}
     sim(q,d) = sim(\text{KG-CQR}(q),d)
     \\
     = <\mathbf{v}_{fuse(q)}, \mathbf{v}_{d}>
\end{aligned}  
\end{equation}

%% file: sections/04_Experiment.tex
\section{Experiment}
\subsection{Experimental Setup}
\textbf{Baseline}: We evaluate our method using three baseline models that encompass diverse document retrieval strategies: (i) BM25 \cite{INR-019}, a classical sparse retrieval model; (ii) DPR \cite{KarpukhinOMLWEC20}, a dense retrieval approach based on a dual-encoder architecture that independently encodes queries and passages, optimizing their embeddings via contrastive loss; and (iii) BGE \cite{bge_embedding}, which combines dense, sparse, and multi-vector retrieval using a self-knowledge distillation framework. To comprehensively examine the impact of KG-CQR on retrieval performance, we further compare KG-CQR with two representative approaches in this research field: Query Expansion \cite{ChenCHW024} and HyDE \cite{GaoMLC23}.
\\
\textbf{Benchmark Datasets}: We evaluate our method on two recent and widely used benchmark datasets: (i) RAGBench \cite{friel2024ragbench}, which spans five distinct industry-specific domains. We use its test set comprising approximately 11,000 instances for retrieval evaluation; and (ii) Multihop-RAG \cite{tang2024multihop}, which includes a knowledge base, a large set of multi-hop queries, corresponding ground-truth answers, and supporting evidence, totaling 2,556 queries for evaluation. For each dataset, the corresponding KG is constructed in three steps, as outlined in Section~\ref{KG-construction}, using the LLaMA-3.3-70B model.
\begin{table*}
  \begin{adjustbox}{max width=\textwidth}
  \begin{tabular}{lcccccccc}
    \toprule
      &\multicolumn{4}{c}{\textbf{RAGBench}}& \multicolumn{4}{c}{\textbf{MultiHop-RAG}} \\
    Model & mAP& Recall@5 &Recall@10 &Recall@25 &mAP &Recall@5&Recall@10 &Recall@25\\
    \midrule
    BM25 & 0.329 & 0.337 & 0.399 & 0.462  & 0.241 &  0.261& 0.353& 0.486 \\
    DPR &0.276& 0.286 & 0.348 & 0.425  & 0.099 &  0.125& 0.183& 0.284\\
   BGE & 0.521&0.510 & 0.589   & 0.657  & 0.227  & 0.251 & 0.357& 0.520\\
     \midrule
    QE + BM25& 0.280 & 0.291 & 0.349 & 0.415 & 0.124 & 0.135 & 0.187 & 0.256 \\
    QE + DPR& 0.251 & 0.260 & 0.317 & 0.392 & 0.058 & 0.069 & 0.101 & 0.169 \\
    QE + BGE & 0.487 & 0.476 & 0.553 & 0.618 & 0.139 & 0.147 & 0.211 & 0.313 \\
   \midrule
    HyDE + DPR & 0.286 & 0.293 & 0.354 & 0.426 & 0.106  & 0.127 & 0.188 & 0.297\\
   HyDE + BGE & 0.516 & 0.507 & 0.586 & 0.638 & 0.232 & 0.256& 0.363& 0.524\\
     \midrule
   KG-CQR + BM25& 0.398& 0.398 & 0.454 & 0.514&  \textbf{0.250}& \textbf{0.267} & \textbf{0.372}& \textbf{0.532} \\
   KG-CQR + DPR & 0.316& 0.319 & 0.384&0.462 & 0.129 & 0.157 & 0.224& 0.340\\
   KG-CQR + BGE& \textbf{0.542}& \textbf{0.529} & \textbf{0.610} & \textbf{0.675} & 0.240& 0.261& 0.371 & 0.525\\  
  \bottomrule
\end{tabular}
\end{adjustbox}
  \caption{Retrieval performance on the RAGBench and MultiHop-RAG datasets with LLaMA-3.3-70B as the backbone LLM for contextual generation}
  \label{tab:freq}
\end{table*}
\subsection{Main Results}
Table~\ref{tab:freq} presents the evaluation results of the retrieval process on both datasets. 
Retrieval accuracy is evaluated using standard metrics such as mean Average Precision (mAP) and Recall@k, where $k \in \{5, 10, 25\}$. The reported results use $\alpha=0.7$ (Equation~\ref{alpha_fuse}), which was found to yield the best performance (the selection of this value is further discussed in Appendix~\ref{alpha-hyper}).
From the results, we draw the following observations:
\\
i) \textbf{Retrieval Performance}:
KG-CQR significantly improves retrieval performance across various retrieval backbones. On the RAGBench dataset, KG-CQR + BGE achieves the best performance overall, with an mAP of 0.542 and Recall@25 of 0.675, outperforming both the baseline models and the HyDE-enhanced variants. On the more challenging MultiHop-RAG dataset, KG-CQR + BM25 achieves the highest recall metrics (e.g., Recall@25 = 0.532), demonstrating KG-CQR’s effectiveness compared to traditional methods.
\\
ii) \textbf{Impact of Query Expansion (QE)}:
Compared with their respective baselines, QE-augmented models generally underperform across both datasets. For instance, QE + BM25 (mAP = 0.280 on RAGBench, 0.124 on MultiHop-RAG) performs notably worse than plain BM25, and similar degradations are observed for DPR and BGE backbones. This suggests that naive query expansion often introduces noise and semantic drift, which outweighs the potential benefits of richer lexical coverage. In contrast, KG-CQR achieves consistent improvements by leveraging structured knowledge for contextually grounded reformulations instead of unguided expansions.
\\
iii) \textbf{Contextual Accuracy}:
The comparatively lower performance of HyDE relative to its baselines indicates potential limitations in relying extensively on synthetic queries generated by LLMs. Specifically, while HyDE offers a straightforward method for enhancing contextual understanding, its effectiveness is notably sensitive to the contextual reliability of the generated content. This highlights the constraints of inadequately grounded synthetic information in retrieval tasks.
\\
iv) \textbf{Diverse Benchmarks}:
Although models like BGE perform well on relatively straightforward datasets such as RAGBench, more complex datasets like MultiHop-RAG demand advanced reasoning capabilities. KG-CQR demonstrates robustness in such settings by effectively handling multi-hop reasoning and maintaining strong performance. These results highlight the importance of retrieval frameworks that integrate contextual understanding and structured knowledge to perform consistently across diverse and complex benchmarks.
\subsection{Detailed Analysis}

\subsubsection{Impact of LLM Backbone}
Table~\ref{tab:across_backbone} illustrates the retrieval performance of KG-CQR when paired with different sizes of language models, using BGE as the underlying retrieval method. 
\begin{table*}
  \begin{adjustbox}{max width=\textwidth}
  \begin{tabular}{lcccccccc}
    \toprule
      &\multicolumn{4}{c}{\textbf{RAGBench}}& \multicolumn{4}{c}{\textbf{MultiHop-RAG}} \\
    Backbone & mAP& Recall@5 &Recall@10 &Recall@25 &mAP &Recall@5&Recall@10 &Recall@25\\
    \midrule
    LLaMA-3.2-3B & 0.537& 0.524 & 0.604 & 0.672 & 0.230 &  0.251 & 0.359 & 0.520 \\
    \midrule
     LLaMA-3.1-8B & 0.538& 0.526 & 0.606 & 0.672 & 0.235 & 0.255 & 0.370 & 0.522 \\
    \midrule
   LLaMA-3.3-70B &0.542& 0.529 & 0.610 & 0.675 & 0.240& 0.261& 0.371 & 0.525\\
  \bottomrule
\end{tabular}
\end{adjustbox}
  \caption{The performance of KG-CQR across various parameter sizes of the backbone LLMs}
  \label{tab:across_backbone}
\end{table*}
Specifically, the LLaMA-3.3-70B model achieves the highest performance across nearly all metrics; however, the performance differences between the 8B and 70B variants are relatively modest, suggesting diminishing returns as model size increases. 
These findings indicate that while larger models do offer performance advantages, KG-CQR remains effective even with relatively smaller backbones such as LLaMA-3.2-3B and LLaMA-3.1-8B. This highlights KG-CQR's practicality for resource-constrained environments, offering a favorable trade-off between retrieval performance and computational cost.

\subsubsection{Ablation Study}
Table~\ref{tab:Ablation} presents an ablation study evaluating the contribution of two core components of KG-CQR: the Textual Triplet Representation (TTR) for extracting subgraph and the Subgraph Completion (Sub.Comp.). 
\begin{table}[!ht]
   \begin{adjustbox}{width=\columnwidth}
    \begin{tabular}{lccc}
      \toprule
     Method  &  Recall@5  & Recall@10 & Recall@25  \\
      \midrule
      W/O TTR &  0.486 & 0.572 & 0.641  \\
      W/O Sub.Comp.  & 0.525 & 0.605 & 0.671  \\
 \midrule
  KG-CQR & 0.529 & 0.610 & 0.675  \\
      \bottomrule
    \end{tabular}
    \end{adjustbox}
   \caption{Ablation study of KG-CQR components on RAGBench}
  \label{tab:Ablation}
\end{table}
As shown in the results, removing TTR (Equation \ref{triplet_embeddings}) leads to the most pronounced drop in performance (e.g., Recall@25 decreases from 0.675 to 0.641), underscoring the importance of TTR in accurately extracting relevant subgraphs that preserve semantic alignment with the query. This confirms that converting structured KG information into textual form plays a critical role in aligning the knowledge with the retrieval task. Similarly, omitting the Subgraph Completion module also results in a notable performance degradation, though less severe than removing TTR. This suggests that while the initial subgraph extraction is vital, enriching the subgraph context via completion further improves the model’s ability to retrieve relevant documents.

\subsubsection{Multi-Step Retrieval for RAG Task}
We evaluate the effectiveness of KG-CQR in multi-step reasoning RAG tasks by integrating its retrieval outputs into the IRCoT framework \cite{TrivediBKS23}. 
\begin{table}[!ht]
   \begin{adjustbox}{width=\columnwidth}
    \begin{tabular}{llccc}
      \toprule
     Model  & Retrieval &F1 $\uparrow$  & Iter $\downarrow$ & Score $\uparrow$ \\
     \midrule
     \multicolumn{5}{c}{\textbf{RAGBench}}  \\
      \midrule
      LLaMA-3.2-3B  &  &0.372 & 3.293 & 3.122 \\
      LLaMA-3.1-8B & BM25 & 0.393 & 2.748 & 3.424 \\
      LLaMA-3.3-70B &  &0.431 & 1.912& 3.449 \\
 \midrule
LLaMA-3.2-3B & \multirow{ 3}{*}{\makecell{KG-CQR\\+BM25}}&0.407 & 2.714& 3.317\\
  LLaMA-3.1-8B &   & 0.410 & 1.834 & 3.603 \\
      LLaMA-3.3-70B &  & 0.443 & 1.393 & 3.826 \\
      \midrule
     \multicolumn{5}{c}{\textbf{HotpotQA}}  \\
      \midrule
      LLaMA-3.2-3B  &  &0.613 & 2.586 & 3.828 \\
      LLaMA-3.1-8B & BM25 & 0.662 & 2.591 & 3.955 \\
      LLaMA-3.3-70B &  &0.663 & 1.465& 4.103 \\
 \midrule
LLaMA-3.2-3B & \multirow{ 3}{*}{\makecell{KG-CQR\\+BM25}}&0.648 & 2.450& 4.106\\
  LLaMA-3.1-8B &   & 0.673 & 2.350 & 4.245 \\
      LLaMA-3.3-70B &  & 0.700 & 1.280 & 4.278 \\
      \midrule
     \multicolumn{5}{c}{\textbf{MuSiQue}}  \\
      \midrule
      LLaMA-3.2-3B &  &0.096 & 3.789 & 1.929 \\
      LLaMA-3.1-8B & BM25  &0.141 & 3.52 & 2.564  \\
      LLaMA-3.3-70B &  &0.374 &2.042& 3.206  \\
 \midrule
     LLaMA-3.2-3B  & \multirow{ 3}{*}{\makecell{KG-CQR\\+BM25}}  &0.124 & 3.245 & 2.334 \\
      LLaMA-3.1-8B &  & 0.223 & 3.089 & 2.783 \\
      LLaMA-3.3-70B &  &0.489 & 2.150& 3.778 \\

      \bottomrule
    \end{tabular}
    \end{adjustbox}
   \caption{Multi-step reasoning RAG performance across various datasets}
  \label{tab:RAG}
\end{table}
To assess the generalizability of KG-CQR, experiments were conducted with three LLMs of varying sizes across multiple datasets. The evaluation highlights the role of KG-CQR in enhancing retrieval performance for reasoning-intensive RAG tasks. We randomly sampled 500 examples from the RAGBench test set and evaluated results using F1, GPT-Score ($Score$) \cite{FuNJ024}, and the average number of reasoning steps ($Iter$). In addition, two widely used multi-hop QA benchmarks, HotpotQA \cite{Yang0ZBCSM18} and MuSiQue \cite{TrivediBKS22}, were included in the evaluation. GPT-Score was computed using GPT-4o through the OpenAI API, based on its performance on the Judge LLM leaderboard\footnote{https://huggingface.co/spaces/AtlaAI/judge-arena}.
As shown in Table~\ref{tab:RAG}, several key insights can be drawn:
i) \textbf{KG-CQR substantially improves retrieval quality across datasets}:
On RAGBench, KG-CQR + BM25 consistently outperforms BM25, with performance gains across all LLM sizes (e.g., F1 improves from 0.393 to 0.410 on LLaMA-3.1-8B). Similar improvements are observed on HotpotQA, where KG-CQR yields a significant gain for the largest model (F1 = 0.700 vs. 0.663). The effect is most pronounced on MuSiQue, where KG-CQR + BM25 achieves F1 = 0.489 with LLaMA-3.3-70B compared to 0.374 for BM25, underscoring its effectiveness for complex multi-hop reasoning (results with the BGE retriever are provided in Appendix~\ref{RAG-BGE}).
ii) \textbf{Contextualized reformulations reduce reasoning iterations}:
KG-CQR consistently decreases the average number of reasoning steps. For example, on HotpotQA with LLaMA-3.3-70B, the number of steps is reduced from 1.465 to 1.280. This suggests that knowledge-grounded query reformulations provide more accurate intermediate evidence, enabling models to converge on answers with fewer redundant reasoning cycles.
iii) \textbf{Cross-model scalability and robustness}:
Performance gains are observed across different LLM sizes, highlighting the adaptability of KG-CQR. Notably, the improvements are more pronounced on datasets requiring deeper reasoning (e.g., RAGBench and MuSiQue), indicating that KG-CQR effectively complements LLM reasoning by supplying better-targeted retrieval contexts.

\subsubsection{Retrieval Latency}
Figure~\ref{fig:latency} compares the relative retrieval latency of the baseline HyDE with three KG-CQR variants: 
\begin{figure}[!t]
    \centering
    \includegraphics[width=\linewidth]{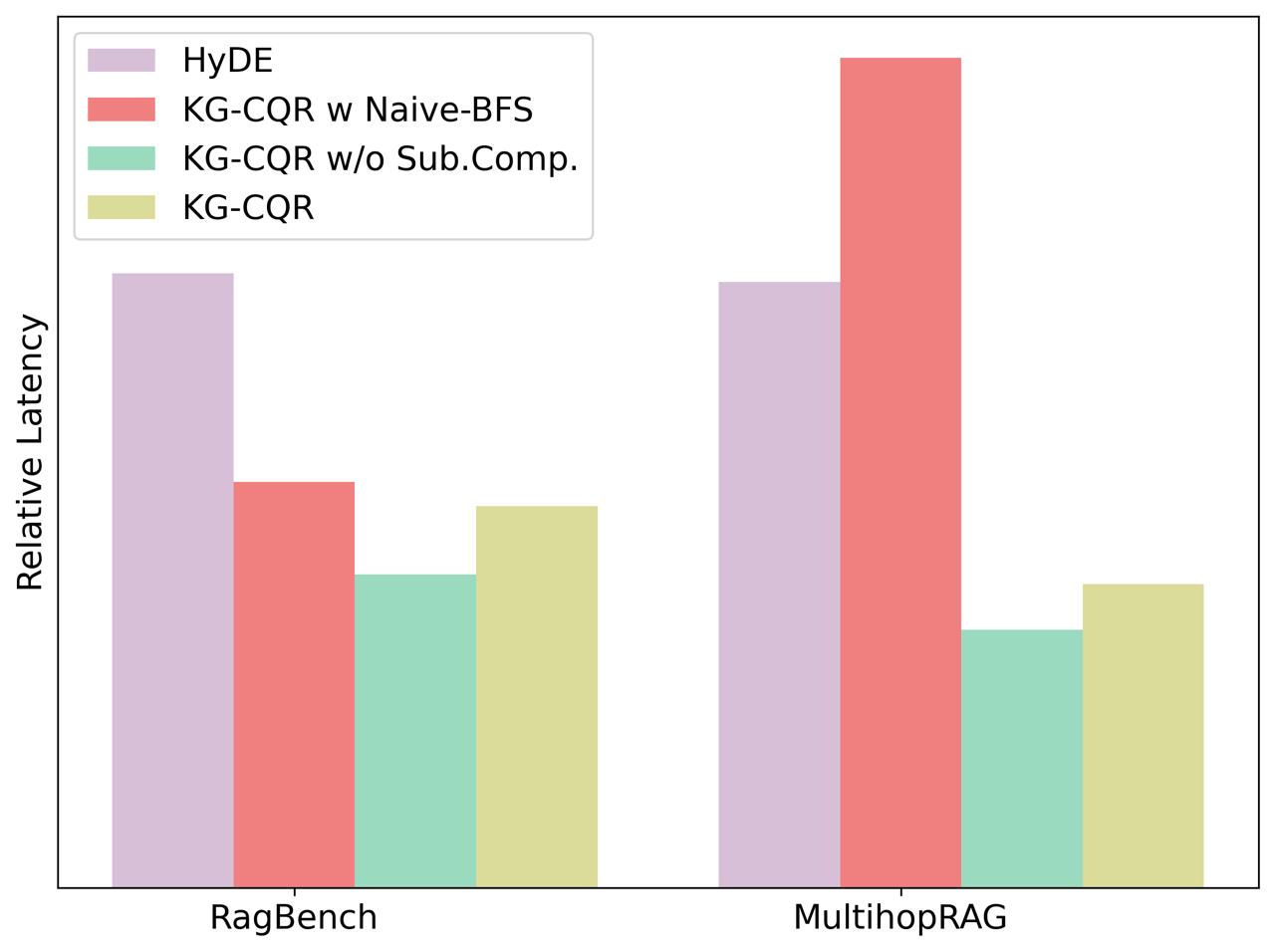}
    \caption{Retrieval latency}
    \label{fig:latency}
\end{figure}
i) \textbf{KG-CQR w/ Naive-BFS)}: use basic BFS algorithm for subgraph completion; ii) \textbf{KG-CQR w/o Sub.Comp.}: removes the subgraph completion module entirely; iii) \textbf{KG-CQR(ours)}: utilizes heuristic-guided Beam Search for more efficient subgraph completion.
The analysis confirms that the proposed KG-CQR with Beam Search strikes an optimal balance between retrieval efficiency and reasoning capability. While KG-CQR without subgraph completion is the fastest, KG-CQR with Beam Search provides a more scalable and semantically expressive alternative with only modest additional cost. In contrast, HyDE and naive BFS approaches incur higher latency, making them less favorable for real-time or large-scale applications.

\subsubsection{Complementarity with Other Methods}
While KG-based methods such as GraphRAG \cite{abs-2404-16130} and HippoRAG \cite{GutierrezS0Y024} emphasize corpus-centric expansion, KG-CQR focuses on query-centric reformulation. To assess their complementarity, we integrated KG-CQR with HippoRAG2 \cite{abs-2502-14802}, as reported in Table~\ref{tab:Integration}.
\begin{table}[!ht]
   \begin{adjustbox}{width=\columnwidth}
    \begin{tabular}{llccc}
      \toprule
     Retrieval & mAP & Recall@5  & Recall@10 & Recall@25 \\
     \midrule
      \makecell{BGE} & 0.221 &0.249 & 0.304 & 0.410 \\
       \midrule
     \makecell{KG-CQR\\ +BGE}  & 0.248 &0.277 & 0.343 & 0.439\\
      \bottomrule
    \end{tabular}
    \end{adjustbox}
   \caption{Retrieval performance of KG-CQR integrated with HippoRAG2 on MultiHop-QA dataset}
  \label{tab:Integration}
\end{table}
The integration yields consistent improvements (e.g., mAP +0.027, Recall@25 +0.029), showing that KG-CQR complements corpus-centric approaches by aligning queries more effectively with relevant evidence. The observed improvements suggest that combining query-centric and corpus-centric KG-based techniques yields a more comprehensive retrieval framework, capable of strengthening both contextual grounding and coverage in multi-hop QA tasks.

%% file: sections/05_Conclusion.tex
\section{Conclusion}
This study presented KG-CQR, a novel retrieval framework that leverages knowledge graphs to enhance contextual query retrieval in RAG systems. By combining subgraph extraction and completion with structured relation representations, KG-CQR enriches query semantics and improves alignment with document embeddings. Experiments on RAGBench and MultiHop-RAG show consistent gains in retrieval performance, while analyses highlight the critical role of textual triplet representation and subgraph completion. Further evaluations on multi-step reasoning RAG tasks indicate improved accuracy while reducing redundant reasoning steps.


%% file: sections/Limitations.tex
\section*{Limitations}
Although KG-CQR demonstrates promising results, several limitations warrant consideration for future improvements:
\\
\textbf{KG Construction Challenges}: The construction of the corpus-centric knowledge graph relies heavily on external LLMs, such as LLaMA-3.3-70B, for entity and relation extraction. This process is susceptible to errors in named entity recognition (NER), relation extraction (RE), and entity linking (EL), which can propagate through the pipeline and affect the quality of the extracted subgraph. In domains with sparse or noisy unstructured data, the resulting KG may lack completeness or accuracy, limiting the effectiveness of KG-CQR.
\\
\textbf{Scalability of Subgraph Extraction}: The subgraph extraction process, while effective, can be computationally intensive for large-scale knowledge graphs with millions of triples. Sentence-level semantic similarity computation with textual triplet representations (TTRs) increases computational overhead, potentially limiting scalability in real-time retrieval systems or resource-constrained environments.
\\
\textbf{Limited Evaluation Scope}: The current evaluation of KG-CQR is restricted to several benchmark datasets. While these datasets are diverse, they may not fully reflect the range and complexity of real-world retrieval scenarios. To more rigorously assess the generalizability of the proposed framework, future work should include evaluations on additional datasets, particularly those that involve cross-lingual settings or highly domain-specific knowledge.


%% file: sections/06_Appendix.tex
\section{Appendix}
\label{sec:appendix}

\subsection{GPT-score Criteria}
\label{GPT-Score}
Following the work in \cite{FuNJ024}, we define the GPT-Score with three  criteria for the measurement as follows:
\begin{itemize}
    \item \textbf{Correctness}: alignment of the generated answer with the reference answer
    \item \textbf{Faithfulness}: whether the generated answer remains true to the given context
    \item \textbf{Relevance}: how well the retrieved context and the generated answer address the query
\end{itemize}
\subsection{Comprehensive Experimential Results}
\subsubsection{Multi-Step Retrieval for RAG with BGE}
\label{RAG-BGE}
Building on the earlier analysis (Table \ref{tab:RAG}), Table \ref{tab:RAG-BGE} presents results for multi-step reasoning RAG performance using BGE as the retrieval baseline, along with KG-CQR. The key observations are as follows:

\begin{table}[!ht]
   \begin{adjustbox}{width=\columnwidth}
    \begin{tabular}{llccc}
      \toprule
     Model  & Retrieval &F1 $\uparrow$  & Iter $\downarrow$ & Score $\uparrow$ \\
     \midrule
     \multicolumn{5}{c}{\textbf{RAGBench}}  \\
      \midrule
      LLaMA-3.2-3B  &  &0.411 & 2.665 & 3.242 \\
      LLaMA-3.1-8B & BGE & 0.434 & 2.272 & 3.528 \\
      LLaMA-3.3-70B &  &0.448 & 1.480& 3.576\\
 \midrule
LLaMA-3.2-3B & \multirow{ 3}{*}{\makecell{KG-CQR\\+BGE}}&0.432 & 2.378& 3.317\\
  LLaMA-3.1-8B &   & 0.438 &1.812& 3.532\\
      LLaMA-3.3-70B &  & 0.452 & 1.230 &3.878\\
      \midrule
     \multicolumn{5}{c}{\textbf{HotpotQA}}  \\
      \midrule
      LLaMA-3.2-3B  &  &0.639 & 2.349 & 3.927 \\
      LLaMA-3.1-8B & BM25 & 0.670 & 2.377 & 4.113 \\
      LLaMA-3.3-70B &  &0.675 & 1.402& 4.222 \\
 \midrule
LLaMA-3.2-3B & \multirow{ 3}{*}{\makecell{KG-CQR\\+BM25}}&0.652 & 2.2240& 4.170\\
  LLaMA-3.1-8B &   & 0.688 & 2.188 & 4.278 \\
      LLaMA-3.3-70B &  & 0.725 & 1.152 & 4.320 \\
      \midrule
     \multicolumn{5}{c}{\textbf{MuSiQue}}  \\
      \midrule
      LLaMA-3.2-3B &  &0.143 & 3.215 & 2.206 \\
      LLaMA-3.1-8B & BM25  &0.203 & 3.050 & 2.679 \\
      LLaMA-3.3-70B &  &0.479 &1.015& 3.908  \\
 \midrule
     LLaMA-3.2-3B  & \multirow{3}{*}{\makecell{KG-CQR\\+BM25}}  &0.175 & 3.066 &2.532 \\
      LLaMA-3.1-8B &  & 0.237 & 2.868 & 2.851 \\
      LLaMA-3.3-70B &  &0.507& 1.936& 3.963\\

      \bottomrule
    \end{tabular}
    \end{adjustbox}
   \caption{Multi-step reasoning RAG performance across various datasets}
  \label{tab:RAG-BGE}
\end{table}

i) \textbf{Dense retrieval outperforms sparse retrieval across all model sizes}: BGE consistently outperforms BM25 in terms of F1 score and GPT-Score, which demonstrates that dense retrieval via BGE retrieves more semantically relevant contexts than BM25, supporting more accurate and efficient reasoning; 
ii) \textbf{KG-CQR improves both BM25 and BGE retrieval}: Adding KG-CQR on top of both BM25 and BGE enhances performance by enriching the query with context-relevant knowledge. Although the improvement margin is narrower in the BGE setting, KG-CQR still consistently enhances performance, highlighting its generality across retrieval methods.
\subsubsection{Fusion Embeddings Experiments}
\label{alpha-hyper}
Table \ref{tab:alpha} shows the comprehensive evaluation on the value of $\alpha$ to fuse the input query and context embeddings (Equation \ref{alpha_fuse}).
\begin{table*}[!ht]
  \begin{adjustbox}{max width=\textwidth}
  \begin{tabular}{lcccccccc}
      \toprule
      &\multicolumn{4}{c}{\textbf{RAGBench}}& \multicolumn{4}{c}{\textbf{MultiHop-RAG}} \\
    Backbone & mAP& Recall@5 &Recall@10 &Recall@25 &mAP &Recall@5&Recall@10 &Recall@25\\
    \midrule
  \multicolumn{9}{c}{$\alpha$ = 0.3} \\
  \hline
   KG-CQR + DPR & 0.320 & 0.325 & 0.385 & 0.462 & 0.143 & 0.169 & 0.239 & 0.354\\
   KG-CQR + BGE& 0.528& 0.513 & 0.596 & 0.664 & 0.224& 0.247& 0.350 & 0.499\\
     \midrule
    \multicolumn{9}{c}{$\alpha$ = 0.5} \\
    \hline
   KG-CQR + DPR & 0.323& 0.327 & 0.391&0.469 & 0.140 & 0.165 & 0.237& 0.351\\
   KG-CQR + BGE& 0.539& 0.527 & 0.609 & \textbf{0.676} & 0.235& 0.253& 0.364 & 0.515\\
     \midrule
  \multicolumn{9}{c}{$\alpha$ = 0.7} \\
  \hline
   KG-CQR + DPR & 0.316& 0.319 & 0.384&0.462 & 0.129 & 0.157 & 0.224& 0.340\\
   KG-CQR + BGE& \textbf{0.542}& \textbf{0.529} & \textbf{0.610} & 0.675 & \textbf{0.240}& \textbf{0.261}& \textbf{0.371} & \textbf{0.525}\\
   \hline
  \bottomrule
\end{tabular}
\end{adjustbox}
  \caption{Fusion embedding results of LLaMA-3.3-70B under different settings of $\alpha$}
  \label{tab:alpha}
\end{table*}
As results, setting $\alpha = 0.7$ consistently yields the best overall performance.

Sequentially, Table \ref{tab:alpha-LLama3.23b} and Table \ref{tab:alpha-LLama3.18b} demonstrate the full experimental results across various backbones, including LLaMA-3.2-3 B and LLaMA-3.1-8B, respectively. Similar to the results on LLaMA-3.3-70B, the KG-CQR + BGE backbone at $\alpha$ = 0.7 yields the best performance for both models, in which LLaMA-3.1-8B shows slight improvements over LLaMA-3.2-3B, particularly in MultiHop-RAG tasks.

\begin{table*}[!ht]
  \begin{adjustbox}{max width=\textwidth}
  \begin{tabular}{lcccccccc}
      \toprule
      &\multicolumn{4}{c}{\textbf{RAGBench}}& \multicolumn{4}{c}{\textbf{MultiHop-RAG}} \\
    Backbone & mAP& Recall@5 &Recall@10 &Recall@25 &mAP &Recall@5&Recall@10 &Recall@25\\
    \midrule
  \multicolumn{9}{c}{$\alpha$ = 0.3} \\
  \hline
   KG-CQR + DPR & 0.312 & 0.319 & 0.382 & 0.458 & 0.129 & 0.152 & 0.221 & 0.337\\
   KG-CQR + BGE& 0.517 & 0.505 & 0.588 & 0.661 & 0.203 & 0.225 & 0.332 & 0.481\\
     \midrule
    \multicolumn{9}{c}{$\alpha$ = 0.5} \\
    \hline
   KG-CQR + DPR & 0.319 & 0.327 & 0.388 & 0.465 & 0.132 & 0.156 & 0.225 & 0.341\\
   KG-CQR + BGE& 0.531 & 0.520 & 0.602 & 0.669 & 0.219 & 0.239 & 0.350 & 0.507\\
     \midrule
  \multicolumn{9}{c}{$\alpha$ = 0.7} \\
  \hline
   KG-CQR + DPR & 0.313 & 0.319 & 0.384 & 0.460 & 0.125 & 0.151 & 0.219& 0.335\\
   KG-CQR + BGE& \textbf{0.537}& \textbf{0.524} & \textbf{0.604} & \textbf{0.672} & \textbf{0.230}& \textbf{0.251}& \textbf{0.366} & \textbf{0.522}\\
  \bottomrule
\end{tabular}
\end{adjustbox}
  \caption{Fusion embedding results of LLaMA-3.2-3B under different settings of $\alpha$}
  \label{tab:alpha-LLama3.23b}
\end{table*}

\begin{table*}[!ht]
  \begin{adjustbox}{max width=\textwidth}
  \begin{tabular}{lcccccccc}
      \toprule
      &\multicolumn{4}{c}{\textbf{RAGBench}}& \multicolumn{4}{c}{\textbf{MultiHop-RAG}} \\
    Backbone & mAP& Recall@5 &Recall@10 &Recall@25 &mAP &Recall@5&Recall@10 &Recall@25\\
    \midrule
  \multicolumn{9}{c}{$\alpha$ = 0.3} \\
  \hline
   KG-CQR + DPR & 0.325 & 0.329 & 0.391 & 0.462 & 0.138 & 0.166 & 0.234 & 0.352\\
   KG-CQR + BGE& 0.523 & 0.509 & 0.591 & 0.659 & 0.216 & 0.237 & 0.341 & 0.489\\
     \midrule
    \multicolumn{9}{c}{$\alpha$ = 0.5} \\
    \hline
   KG-CQR + DPR & 0.327 & 0.330 & 0.394 & 0.467 & 0.136 & 0.162 & 0.233 & 0.351\\
   KG-CQR + BGE& 0.535 & 0.522 & 0.603 & 0.669 & 0.227 & 0.247 & 0.359 & 0.510\\
     \midrule
  \multicolumn{9}{c}{$\alpha$ = 0.7} \\
  \hline
   KG-CQR + DPR & 0.318 & 0.322 & 0.387 & 0.462 & 0.127 & 0.151 & 0.220& 0.338\\
   KG-CQR + BGE& \textbf{0.538}& \textbf{0.526} & \textbf{0.606} & \textbf{0.672} & \textbf{0.236}& \textbf{0.255}& \textbf{0.370} & \textbf{0.522}\\
  \bottomrule
\end{tabular}
\end{adjustbox}
  \caption{Fusion embedding results of LLaMA-3.1-8B under different settings of $\alpha$}
  \label{tab:alpha-LLama3.18b}
\end{table*}
\subsection{BFS with Beam Search Algorithm}
\label{BFS-section}
Algorithm \ref{bfs_paths} presents the pseudocode for the BFS with Beam Search. Given the hyperparameter Beam width (e.g., equal to 3), the algorithm explores explicit paths (triplets) that represent meaningful connections between entities within the given subgraph.
\begin{algorithm}[!ht]
\caption{BFS Algorithm with Beam Search}\label{bfs_paths}
\begin{algorithmic}[1]
\Function{BFSBeam}{$\mathcal{T}_{KG}$, $e_s$, $e_t$, $T_{\text{set}}$, $L$}

\State $Q \gets \text{Queue}(\{\langle e_s, \emptyset \rangle\})$; $P \gets \emptyset$ 
\State Load Embedding mode: enc
\State $v_q \gets q \neq \emptyset ? \text{enc}(q) : \text{None}$ 
\State $S \gets \emptyset$ 
\State $W \gets 3$ \Comment{Beam width for Beam Search}

\While{$Q \neq \emptyset$}
    \State $(\text{node}, p) \gets Q.\text{dequeue}()$ 
    \If{$|p| > L$} 
        \State \textbf{continue}
    \EndIf
    \If{$\text{node} = e_t$ \textbf{and} $p \neq \emptyset$} 
        \State $P \gets P \cup \{p\}$ 
    \EndIf
    \If{$\text{node} = e_t$}
        \State \textbf{continue}
    \EndIf
    \For{$\{u, r, v, \text{TTR}\} \in \mathcal{T}_{KG}(u, v, r)$ \textbf{where} $u = \text{node}$} 
        \If{$v \notin p.\text{entities}$} 
            \State $v_p \gets \text{enc}(\text{TTR})$ 
            \State $s \gets v_q \neq \text{None} ? \text{Mean}(\cos(v_p, v_q)) : 0$ 
            \State $p_{\text{new}} \gets p \cup \{u, r, v\}$ 
            \State $S \gets S \cup \{(p_{\text{new}}, s)\}$ 
        \EndIf
    \EndFor
    \State Sort $S$ by score descending 
    \For{$(p_{\text{new}}, s) \in S$ \textbf{take top} $W$} 
        \State $u \gets p_{\text{new}}.\text{last\_node}$
        \State $Q.\text{enqueue}((u, p_{\text{new}}))$
    \EndFor
    \State $S \gets \emptyset$ 
\EndWhile

\State \Return $P$

\EndFunction

\end{algorithmic}
\end{algorithm}

\subsection{Error Analysis with Examples}
To better understand the behavior of the KG-CQR, we performed a qualitative error analysis on six representative multi-hop queries from the MultiHop-RAG dataset with three corrected retrievals (Table \ref{tab:correct_samples}) and three with incorrect retrievals (Table \ref{tab:incorrect_samples}). We compared the outputs of KG-CQR against those of HyDE and the human-annotated Ground Truth.

Based on the results in Table \ref{tab:correct_samples}, there are several assumptions as follows: i) KG-CQR demonstrates strong performance in disambiguating entities. For instance, in the query “Did one of CBS’s performers create a scandal?”, KG-CQR retrieves documents specifically related to the mentioned performer and event. This shows that incorporating knowledge graph information improves precision by retrieving documents more closely aligned with the query context; ii) In time-sensitive queries like “Which events occurred in Week 12?”, KG-CQR accurately retrieves temporally relevant content, whereas HyDE often returns general or loosely connected documents. This suggests that KG signals enhance temporal grounding in multi-hop retrieval tasks; iii) For bridge-type queries that require chaining multiple pieces of information (e.g., “Does the article from Wendy refer to the same city?”), KG-CQR performs well by retrieving documents that correctly capture the intermediate and final entities. This indicates improved multi-hop coherence over baseline methods.

Despite these strengths, the proposed KG-CQR shows notable limitations in the following areas (Table \ref{tab:incorrect_samples}): i) \textbf{Contextual Drift and Irrelevant Retrievals}: KG-CQR struggles with queries requiring fine-grained temporal reasoning, comparative analysis, or interpretation of subjective content. These limitations stem from insufficient temporal representation and the lack of deep semantic modeling needed to capture nuanced relationships and contrasting viewpoints; ii) \textbf{Limited Multi-hop Coherence}: For queries requiring reasoning across multiple documents, KG-CQR sometimes retrieved disconnected evidence, failing to form a complete answer path.

\begin{table*}[!ht]
    \centering
    \renewcommand{\arraystretch}{1.2}
    \scriptsize
    \begin{adjustbox}{width=\textwidth}
    \begin{tabular}{p{2cm} p{4cm} p{4cm} p{4cm}}
        \toprule
        \toprule
        \multicolumn{1}{c}{\textbf{Query $q$}} & 
        \multicolumn{1}{c}{\textbf{HyDE@5}} & 
        \multicolumn{1}{c}{\textbf{KG-CQR@5}} & 
        \multicolumn{1}{c}{\textbf{\shortstack{Ground Truth}}} \\
        \midrule
       Did the CBSSports.com article report Kenneth Walker III remaining healthy and uninjured during a game, similarly to how the Sporting News article reports injuries for Tee Higgins, Noah Brown, Treylon Burks, and Kadarius Toney preventing their participation in Week 12?
        & 
        \textbf{\textbf{D1:}} Meanwhile, corner CJ Henderson (concussion) was a full participant on Friday and carries no designation heading into the weekend...'\newline
        \textbf{\textbf{D2:}} He left Week 2 after suffering a concussion and was absent in Week 3; then was not part of the game plan much in Week 4 (7.7\% target share against Philadelphia)...'\newline
        \textbf{\textbf{D3:}} He was ruled questionable to return. NFL Media reported on Monday that Kupp suffered a low ankle sprain and will be evaluated going forward...'\newline
        \textbf{\textbf{D4:}} \textcolor{blue}{Geno Smith's struggles complicate their fantasy prospects, as well...'}\newline
        \textbf{\textbf{D5:}} Head coach Ron Rivera called the injury "significant" earlier this week...'\newline
        &
        \textbf{\textbf{D1:}} \textcolor{blue}{Geno Smith's struggles complicate their fantasy prospects, as well...'}\newline
        \textbf{\textbf{D2:}} Meanwhile, corner CJ Henderson (concussion) was a full participant on Friday and carries no designation heading into the weekend...'\newline
        \textbf{\textbf{D3:}} \textcolor{blue}{When asked if that means for Thursday's matchup against San Francisco, the coach said, "I would think so."...'}\newline
        \textbf{\textbf{D4:}} Week 11 of the 2023 NFL season has provided plenty of drama, from the Bears hanging with the Lions to the Giants getting a rebound from emergency quarterback Tommy DeVito...'\newline
        \textbf{\textbf{D5:}} \textcolor{blue}{Miami ruled him questionable to return with a knee injury, and while he later returned to the sidelines from a locker-room visit, he was replaced on the field indefinitely by Raheem Mostert...'}\newline
        &
        \textbf{\textbf{D1:}} When asked if that means for Thursday's matchup against San Francisco, the coach said, "I would think so."...'\newline
        \textbf{\textbf{D2:}} Geno Smith's struggles complicate their fantasy prospects, as well...'\newline
        \textbf{\textbf{D3:}} Walker's struggled under center, Tillman is set up for a high-usage day against the Rams with Amari Cooper (ribs) banged up...'\newline
        \textbf{\textbf{D4:}} Miami ruled him questionable to return with a knee injury, and while he later returned to the sidelines from a locker-room visit, he was replaced on the field indefinitely by Raheem Mostert...'\newline
        \\
        \midrule
       Does the article from Wired suggest that Sony headphones do not offer the best value in their class during the Walmart Cyber Monday Deals, while the article from Music Business Worldwide indicates that Artists are seeking deals that offer more control and better economics, or do both articles suggest a common trend in seeking value and control in their respective fields?
        & 
        \textbf{\textbf{D1:}} Black Friday is often a boon for deals on headphones and earbuds, and this year is no different...'\newline
        \textbf{\textbf{D2:}} Engadget has been testing and reviewing consumer tech since 2004. Our stories may include affiliate links; if you buy something through a link, we may earn a commission...'\newline
        \textbf{\textbf{D3:}} But in both the case of Universal Music Group and Warner Music Group, they’re – currently anyway – not the biggest megastars on either company’s books...'\newline
        \textbf{\textbf{D4:}} This is one of the few sales we've seen all year, which makes their very high asking price a lot more palatable...'\newline
        \textbf{\textbf{D5:}} Nothing is more frustrating than buying a new pair of headphones, an OLED TV, or a backpack just to find out that you could have gotten it for a lot cheaper somewhere else...'\newline
        &
        \textbf{\textbf{D1:}} \textcolor{blue}{They're light on extras like noise canceling but at this price, they're a great investment as your go-to workout companions...'}\newline
        \textbf{\textbf{D2:}} Black Friday is often a boon for deals on headphones and earbuds, and this year is no different...'\newline
        \textbf{\textbf{D3:}} Luckily they’ve already gotten a discount, which makes it easier to land their class-leading noise canceling, great sound, and luxuriously comfy design that’s loaded with modern features...'\newline
        \textbf{\textbf{D4:}} Engadget has been testing and reviewing consumer tech since 2004. Our stories may include affiliate links; if you buy something through a link, we may earn a commission. Read more about how we evaluate products...'\newline
        \textbf{\textbf{D1:}} This is one of the few sales we've seen all year, which makes their very high asking price a lot more palatable...'\newline
        &
        \textbf{\textbf{D1:}} Spanish and Latin artists have much more options to develop their audiences and monetize their music at each stage of their career...'\newline
        \textbf{\textbf{D2:}} They're light on extras like noise canceling but at this price, they're a great investment as your go-to workout companions...'\newline
        \\
        \midrule
       Which company, covered by both TechCrunch and The Verge, is not only claimed to have developed an AI model with superior architecture that rivals GPT-4 but also has been accused of altering the internet's appearance and harming news publishers' bottom lines through anticompetitive practices?
        & 
        \textbf{\textbf{D1:}} Hey, folks, welcome to Week in Review (WiR), TechCrunch’s regular newsletter that recaps the past few days in tech. AI stole the headlines once again, with tech giants from Google to X (formerly Twitter) heading off against OpenAI for chatbot supremacy...'\newline
        \textbf{\textbf{D2:}} And on a company level, Meta is doing all it can to encourage collaboration and “openness,” recently partnering with Hugging Face to launch a new startup accelerator designed to spur adoption of open source AI models...'\newline
        \textbf{\textbf{D3:}} Google, OpenAI and Microsoft, a close OpenAI partner and investor, have been among the chief critics of Meta’s open source AI approach, arguing that it’s potentially dangerous and disinformation-encouraging...'\newline
        \textbf{\textbf{D4:}} By 2020, the Knowledge Graph had grown to 500 billion facts about 5 billion entities. But much of the “collective intelligence” that Google tapped into was content “misappropriated from Publishers,” the complaint alleges...'\newline
        \textbf{\textbf{D5:}} The lawsuit reiterates this concern, claiming that Google’s recent advances in AI-based search were implemented with “the goal of discouraging end-users from visiting the websites of Class members who are part of the digital news and publishing line of commerce.”...'\newline
        &
        \textbf{\textbf{D1:}} \textcolor{blue}{A new class action lawsuit filed this week in the U.S. District Court in D.C. accuses Google and parent company Alphabet of anticompetitive behavior in violation of U.S. antitrust law, the Sherman Act, and others, on behalf of news publishers...'}\newline
        \textbf{\textbf{D2:}} \textcolor{blue}{This week, Google took the wraps off of Gemini, its new flagship generative AI model meant to power a range of products and services including Bard, Google’s ChatGPT competitor...'}\newline
        \textbf{\textbf{D3:}} The lawsuit reiterates this concern, claiming that Google’s recent advances in AI-based search were implemented with “the goal of discouraging end-users from visiting the websites of Class members who are part of the digital news and publishing line of commerce.”...'\newline
        \textbf{\textbf{D4:}} By 2020, the Knowledge Graph had grown to 500 billion facts about 5 billion entities. But much of the “collective intelligence” that Google tapped into was content “misappropriated from Publishers,” the complaint alleges...'\newline
        \textbf{\textbf{D5:}} Hey, folks, welcome to Week in Review (WiR), TechCrunch’s regular newsletter that recaps the past few days in tech. AI stole the headlines once again, with tech giants from Google to X (formerly Twitter) heading off against OpenAI for chatbot supremacy...'\newline
        &
        \textbf{\textbf{D1:}} “I used to do those types of tactics, so I couldn’t hate on anybody personally,” she said. “If people have a problem with Google’s results, they have to ask themselves, is it the fault of the SEOs?” she asked...'\newline
        \textbf{\textbf{D2:}} A new class action lawsuit filed this week in the U.S. District Court in D.C. accuses Google and parent company Alphabet of anticompetitive behavior in violation of U.S. antitrust law, the Sherman Act, and others, on behalf of news publishers...'\newline
        \textbf{\textbf{D3:}} Sure, she called herself a “thought leader,” and yes, sure, she had changed her last name to improve her personal branding by more closely associating herself with her grandmother’s uncle, the artist Man Ray...'\newline
        \textbf{\textbf{D4:}} This week, Google took the wraps off of Gemini, its new flagship generative AI model meant to power a range of products and services including Bard, Google’s ChatGPT competitor...'\newline
        \\
        \bottomrule
        \bottomrule
    \end{tabular}
    \end{adjustbox}
    \caption{Examples of KG-CQR with correctly retrieved documents. Blue texts are corrected retrieved documents}
    \label{tab:correct_samples}
\end{table*}

\begin{table*}[!ht]
    \centering
    \renewcommand{\arraystretch}{1.2}
    \scriptsize
    \begin{adjustbox}{width=\textwidth}
    \begin{tabular}{p{2cm} p{4cm} p{4cm} p{4cm}}
        \toprule
        \toprule
        \multicolumn{1}{c}{\textbf{Query $q$}} & 
        \multicolumn{1}{c}{\textbf{HyDE@5}} & 
        \multicolumn{1}{c}{\textbf{KG-CQR@5}} & 
        \multicolumn{1}{c}{\textbf{\shortstack{Ground Truth}}} \\
        \midrule
        Has the approach of 
 Sportsbooks in adjusting betting lines and odds, as reported by Sporting News \textcolor{red}{after October 4, 2023, and before November 1, 2023}, remained consistent?
        & 
        \textbf{\textbf{D1:}} For instance, when examining odds for the next Super Bowl champion released shortly after the previous Super Bowl, these odds are based mostly on the recently concluded season...\newline
        \textbf{\textbf{D2:}} They are basing their odds on past performance and expected future accomplishments, as well as the quality of the team around the top candidates for the award. Thus, the odds are quite favorable...\newline
        \textbf{\textbf{D3:}} When such information becomes public, sportsbooks may adjust the odds accordingly. Professional Bettors: Large wagers from sharp bettors or professional gamblers can cause the lines to shift...\newline
        \textbf{\textbf{D4:}} The past few weeks of the 2023 NFL season have reminded us that no matter how smooth you sail to start the voyage, choppy waters will surely come at some point...\newline
        \textbf{\textbf{D5:}} Let’s say the Chiefs win by exactly three, a distinct possibility since a single field goal decides most NFL games...\newline
        &
        \textbf{\textbf{D1:}} For instance, when examining odds for the next Super Bowl champion released shortly after the previous Super Bowl, these odds are based mostly on the recently concluded season...\newline
        \textbf{\textbf{D2:}} When such information becomes public, sportsbooks may adjust the odds accordingly. Professional Bettors: Large wagers from sharp bettors or professional gamblers can cause the lines to shift...\newline
        \textbf{\textbf{D3:}} They are basing their odds on past performance and expected future accomplishments, as well as the quality of the team around the top candidates for the award...\newline
        \textbf{\textbf{D4:}} The past few weeks of the 2023 NFL season have reminded us that no matter how smooth you sail to start the voyage, choppy waters will surely come at some point. We started the first six weeks with a best bets winning percentage of well over ...\newline
        \textbf{\textbf{D5:}} Do point spread odds change?

Yes, point spread odds can change, and these shifts are commonly referred to as "line movement."\newline
        &
        \textbf{\textbf{D1:}} It's important to note that in PGA and other golf tournaments, there are usually many players, so the odds can be much higher than in head-to-head sports matchups, given the broader field of competition...\newline
        \textbf{\textbf{D2:}} BetMGM Sportsbook: As one of the most recognizable names in the gambling industry, BetMGM knows how to attract and keep customers with competitive odds for all bet types, including futures bets and the NBA Rookie of the Year...\newline
        \textbf{\textbf{D3:}} When the lines are first released for NBA ROTY honors, the season hasn’t even started yet, so there are no statistics, trends, or player news...\newline
        \textbf{\textbf{D4:}} Does overtime count in my moneyline bet?

Yes, in most sports and with most sportsbooks (including new betting sites), overtime (or any extra time or tiebreakers) does count in a moneyline bet.\newline
        \\
        \midrule
       Does the TechCrunch article on generative AI in the enterprise suggest that \textcolor{red}{CIOs are more cautious in their AI adoption strategy compared to the belief of business leaders} mentioned in another TechCrunch article, who think AI will be essential for all businesses within five years?
        & 
        \textbf{\textbf{D1:}} To hear the hype from vendors, you would think that enterprise buyers are all in when it comes to generative AI. But like any newer technology, large companies tend to move cautiously...\newline
        \textbf{\textbf{D2:}} I’d venture to guess more exposure for its burgeoning generative AI platform...\newline
        \textbf{\textbf{D3:}} Expect more moves like that from 2024’s OpenAI as the caution and academic reserve that the previous board exerted gives way to an unseemly lust for markets and customers...\newline
        \textbf{\textbf{D4:}} Google, OpenAI and Microsoft, a close OpenAI partner and investor, have been among the chief critics of Meta’s open source AI approach, arguing that it’s potentially dangerous and disinformation-encouraging...\newline
        \textbf{\textbf{D5:}} The NMPA’s submission, dated October 30, 2023, pulls no punches.

It starts off by stressing that its membership – US music publishers major and independent – are “not opposed” to AI...\newline
        &
        \textbf{\textbf{D1:}} To hear the hype from vendors, you would think that enterprise buyers are all in when it comes to generative AI...\newline
        \textbf{\textbf{D2:}} I’d venture to guess more exposure for its burgeoning generative AI platform. IBM’s most recent earnings were boosted by enterprises’ interest in generative AI, but the company has stiff competition in Microsoft and OpenAI...\newline
        \textbf{\textbf{D4:}} Expect more moves like that from 2024’s OpenAI as the caution and academic reserve that the previous board exerted gives way to an unseemly lust for markets and customers...\newline
        \textbf{\textbf{D4:}} The NMPA’s submission, dated October 30, 2023, pulls no punches.

It starts off by stressing that its membership – US music publishers major and independent – are “not opposed” to AI...'\newline
        \textbf{\textbf{D5:}} Google

On generative AI, Google’s report discusses “recent progress in large-scale AI models” which it suggests...
\newline
        &
        \textbf{\textbf{D1:}} “So we’ve been doing this whole push for AI over the last maybe six or nine months and we’re at the point right now where we’re building specific use cases for each different team and function within the firm.”...\newline
        \textbf{\textbf{D2:}} Third, the application is only as sophisticated as the data that it is fed. Proprietary data is necessary for specific and relevant insights and to ensure others cannot replicate the final product...\newline
        \textbf{\textbf{D3:}} That’s going to take setting up some structure and organization around how this gets implemented over time, says Jim Rowan, principal at Deloitte, who is working with clients around how to build generative AI across companies in an organized fashion...\newline
        \\
        \midrule
       Does 'The Independent - Life and Style' article suggesting \textcolor{red}{Prince William's emotional state regarding Princess Diana's death} align with the same publication's depiction of the events leading up to her death in 'The Crown season six'?
        & 
        \textbf{\textbf{D1:}} He is not located, but later walks back to the house on his own accord, drenched in rain. “14 hours, that poor boy was gone,” the Queen later says...\newline
        \textbf{\textbf{D2:}} The show also features the pair’s death in a car crash in Paris.

As the new season arrives, and fans wonder what in The Crown is based in reality, here’s everything you need to know...\newline
        \textbf{\textbf{D3:}} She then poses for them in her swimsuit, but complains in a later episode that they can “never relax” with the press “constantly” around...\newline
        \textbf{\textbf{D4:}} After staying several days on Mohamed Al Fayed’s yacht, the boys return home to London where their father, the then-Prince of Wales, accompanies them to Balmoral Castle to vacation with the rest of the royal family in Scotland...\newline
        \textbf{\textbf{D5:}} During the interview, the outlet noted that Smith said his wife’s memoir “kind of woke him up” and that he has now realised she is more...\newline
        &
        \textbf{\textbf{D1:}} The show also features the pair’s death in a car crash in Paris.

As the new season arrives, and fans wonder what in The Crown is based in reality, here’s everything you need to know...\newline
        \textbf{\textbf{D2:}} He is not located, but later walks back to the house on his own accord, drenched in rain. “14 hours, that poor boy was gone,” the Queen later says...\newline
        \textbf{\textbf{D3:}} After staying several days on Mohamed Al Fayed’s yacht, the boys return home to London where their father, the then-Prince of Wales, accompanies them to Balmoral Castle to vacation with the rest of the royal family in Scotland...\newline
        \textbf{\textbf{D4:}} She then poses for them in her swimsuit, but complains in a later episode that they can “never relax” with the press “constantly” around...\newline
        \textbf{\textbf{D5:}} Asks the Queen if she’d received the invitation to Camilla’s 50th birthday, to which she says she has, but cannot attend as she’s in Derbyshire...\newline
        &
        \textbf{\textbf{D1:}} Stay ahead of the trend in fashion and beyond with our free weekly Lifestyle Edit newsletter Stay ahead of the trend in fashion and beyond with our free weekly Lifestyle Edit newsletter Please enter a valid email address...\newline
        \textbf{\textbf{D2:}} However, at the inquest into the death in 2007, the jury were shown CCTV footage of him purchasing an engagement ring worth £11,600 in a jewellers across the square from the Ritz on the afternoon of the crash...\newline
        \\
        \bottomrule
        \bottomrule
    \end{tabular}
    \end{adjustbox}
    \caption{Examples of KG-CQR with incorrectly retrieved documents. Red texts indicate notable limitations of KG-CQR in several areas, such as contextual drift or limited complex multi-hop coherence}
    \label{tab:incorrect_samples}
\end{table*}

\subsection{Prompt Template}
\label{prompt-template}
\begin{figure*}[!t]
\includegraphics[width=\textwidth]{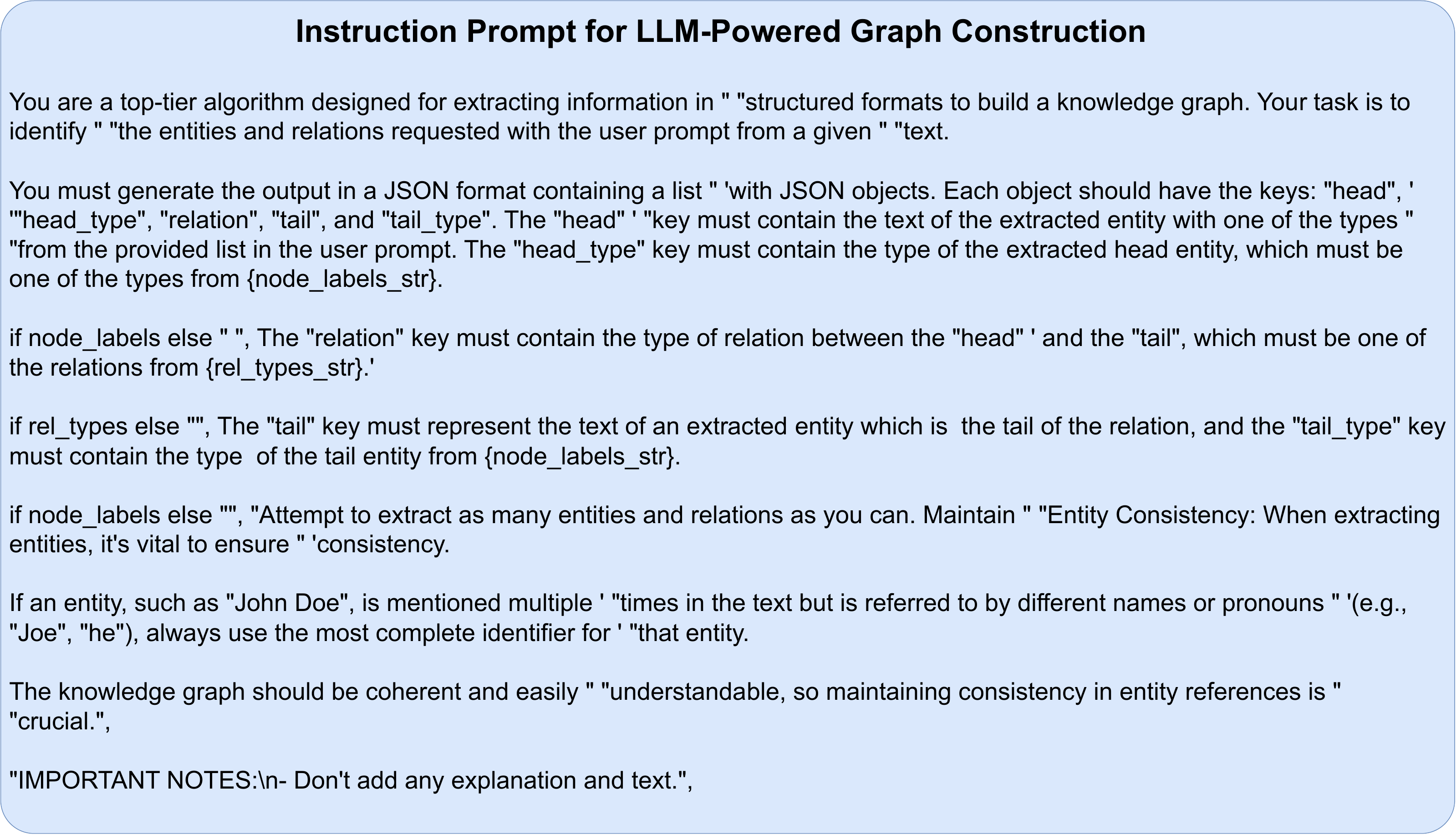}
\caption{Prompt the task instruction for KG construction}
\label{fig:kg_prompt}
\end{figure*}

\begin{figure*}[!t]
\includegraphics[width=\textwidth]{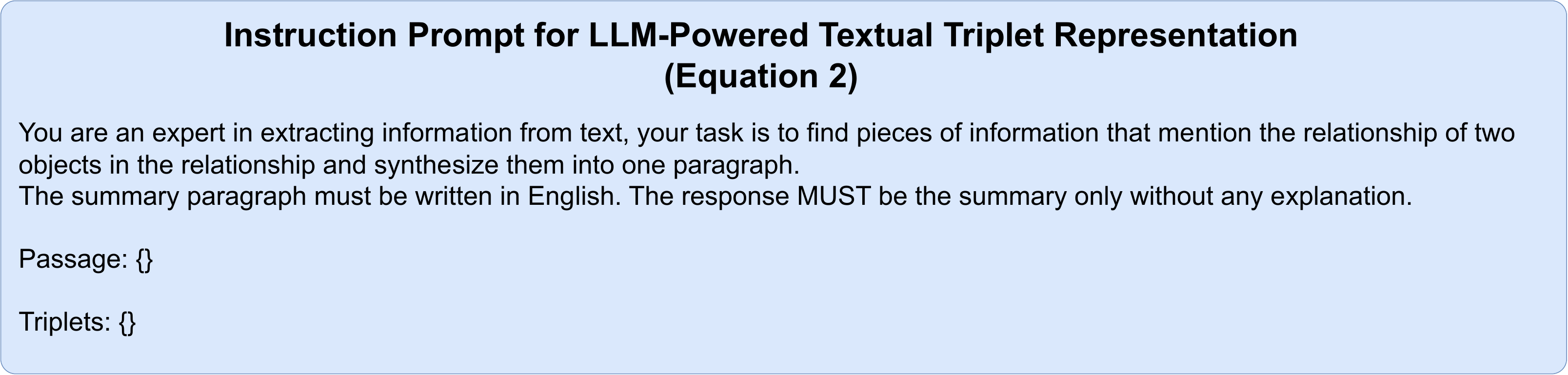}
\caption{Prompt the task instruction for textual triplet representation.}
\label{fig:ttr_prompt}
\end{figure*}

\begin{figure*}[!t]
\includegraphics[width=\textwidth]{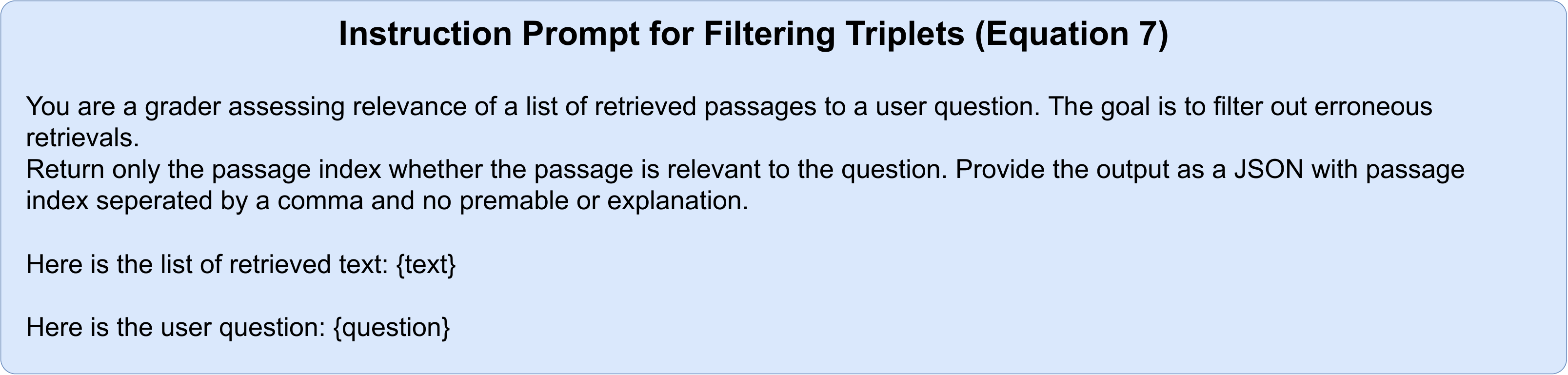}
\caption{Prompt the task instruction for filtering irrelevant triplets}
\label{fig:filter_prompt}
\end{figure*}

\begin{figure*}[!t]
\includegraphics[width=\textwidth]{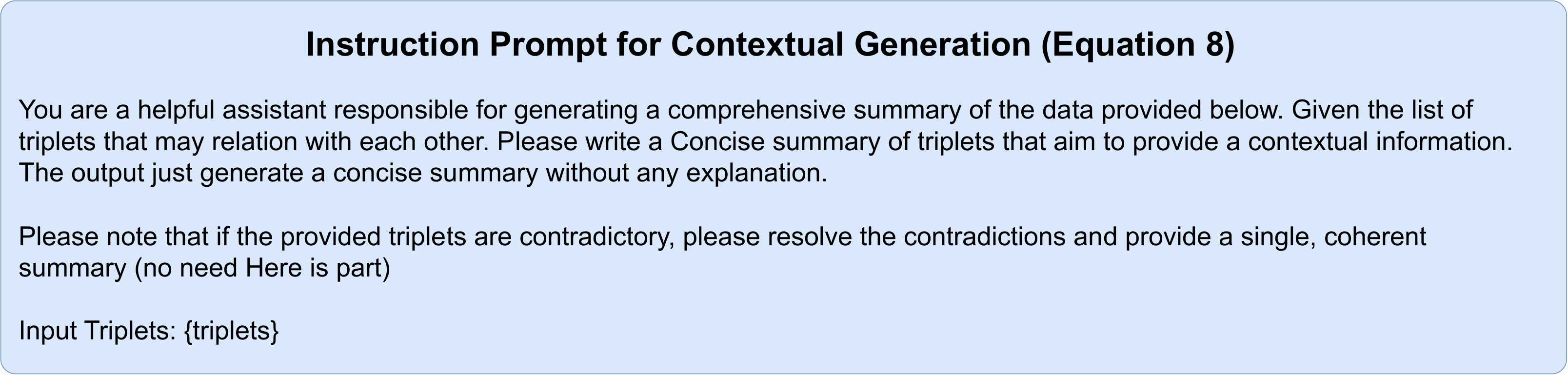}
\caption{Prompt the task instruction for contextual representation}
\label{fig:context-gen_prompt}
\end{figure*}

For better reproducibility, we present all prompt templates in the appendix. Below is a quick reference list outlining the prompt templates and their usages:
\begin{itemize}
    \item Figure \ref{fig:kg_prompt}: Prompt the task instruction for KG construction.
    \item Figure \ref{fig:ttr_prompt}: Prompt the task instruction for textual triplet representation.
    \item Figure \ref{fig:filter_prompt}: Prompt the task instruction for filtering irrelevant triplets.
    \item Figure \ref{fig:context-gen_prompt}: Prompt the task instruction for contextual generation.
\end{itemize}

